**What You Read is What You Classify: Highlighting Attributions to Text and Text-Like Inputs**

*Daniel S. Berman[1], Brian Merritt[1], Stanley Ta[1], Dana Udwin[1], Amanda Ernlund[1], Jeremy Ratcliff[1]^, Vijay Narayan[1] **

1. Johns Hopkins Applied Physics Laboratory, 11100 Johns Hopkins Rd., Laurel, MD 20723, USA.

^ Current affiliation: Google DeepMind, 14–18 Handyside Street, London, N1C 4DN, United Kingdom

* Corresponding author: Vijay.narayan@jhuapl.edu

# Abstract

At present, there are no easily understood explainable artificial intelligence (AI) methods for discrete token inputs, like text. Most explainable AI techniques do not extend well to token sequences, where both local and global features matter, because state-of-the-art models, like transformers, tend to focus on global connections. Therefore, existing explainable AI algorithms fail by (i) identifying disparate tokens of importance, or (ii) assigning a large number of tokens a low value of importance. This method for explainable AI for tokens-based classifiers generalizes a mask-based explainable AI algorithm for images. It starts with an Explainer neural network that is trained to create masks to hide information not relevant for classification. Then, the Hadamard product of the mask and the continuous values of the classifier's embedding layer is taken and passed through the classifier, changing the magnitude of the embedding vector but keeping the orientation unchanged. The Explainer is trained for a taxonomic classifier for nucleotide sequences and it is shown that the masked segments are less relevant to classification than the unmasked ones. This method focused on the importance the token as a whole (i.e., a segment of the input sequence), producing a human-readable explanation.

# Introduction

Trust is an essential component to the adoption and utilization of new artificial intelligence (AI) tools (1). Trust is especially important when AI attempts to replace a non-black box technology but no mechanism is offered for determining accountability or mitigating errors and bias. Therefore, developing explainable AI techniques and methods, especially those that are highly interpretable, is beneficial to user adoption (2,3).

The most advanced explainable AI techniques are typically designed for image classification tasks because the continuous values and ready interpretability of image subsets lend themselves to the two major types of explainable AI techniques. The first of these uses the model's activation, or backpropagated signals, to determine the most important features. Techniques that use the activation signals include the class activation map (CAM) (4) that uses global average pooling; techniques like Grad-CAM(5), Zeigler and Fergus(6), and Wagner et al.(7), that use gradients; or combinations of activations and gradients, like Grad-CAM++ (8), among others. These methods have been most successfully applied to convolutional neural networks and fully connected networks. While convolutional neural networks can be useful for text classification, recurrent neural networks and transformer architectures are more frequently used for text classification because they have substantially more complicated interactions and can produce maps of disconnected, important tokens.

The second major type of explainable AI technique is perturbation-based, where small changes are made to the inputs to identify which feature has the highest impact on the final output. Shapley additive explanations (SHAP)(9) values and local interpretable model-agnostic explanations (LIME)(10) fall into this category. Both of these methods are computationally intensive and incur most of the computational cost at runtime, rather than upfront.

Unlike either of these two major types of explainable AI techniques, the "what-you-see" algorithm conceived by Stalder et al. (11) works by creating a new neural network—the Explainer—that generates a masked for an input image. This mask blocks out all the irrelevant parts of the image and passes that masked image through the original AI model that needs explaining, the Explanandum. This explainable AI method incurs all the computational costs when training the Explainer. Then, an image need only be passed through the trained Explainer to generate an explanation (i.e., the unmasked parts of the image). Additionally, the explainable AI method uses a loss function that values continuous regions, so the masks themselves are easily interpreted.

Neither this technique nor many of the existing explainable AI techniques can be applied outright to tokenized data, such as text or nucleotide/amino acid sequences; for example, perturbing a token isn't meaningful because tokens are categorical. An embedding can be perturbed, but that will reveal more about the components of the embedding vector, not the tokens that preceded the embeddings. The most common explainable AI algorithm for tokenized data is SHAP values(9) and LIME(10), but the results are often difficult to interpret because tokens are evaluated one-at-a-time. Text-based transformer and recurrent neural network architectures, which outperform CNNs based architectures, do not consider local structure of the inputs, so important tokens may be separated by a large distance. Additionally, because the importance values range from -1 to 1, a large number of small positive values can overrule a single large negative value, or visa-versa, obscuring the true importance landscape.

We propose a novel approach to explainable AI for text with our adaptation of Stadler et al.(11) to provide meaningful explanations for classification decisions on tokenized sequences. We demonstrate our method on a genomic taxonomic classification task at three nested levels of classification—superkingdom, phylum and genus—using the Nucleotide Transformer 50 million parameter model (NT50m)(12) and the BERTax dataset(13).

# Methods

The approach to the what-you-read model is very similar to that of the what-you-see model, with two slight modifications (**Figure 1**). The first modification pertains to masking tokens. Because the pixel values in images are continuous, it is trivial to take the Hadamard product of the mask of real values against the image, and pass the result to the Explanandum. The product of the mask and the image is a masked image, where masked values nearer to zero are 'darker'. Because the mask values are continuous, the entire process of passing an input image through the Explainer, then its output through the Explanandum, is differentiable. As a result, the weights of the Explainer can be changed, while the Explanandum weights are frozen.

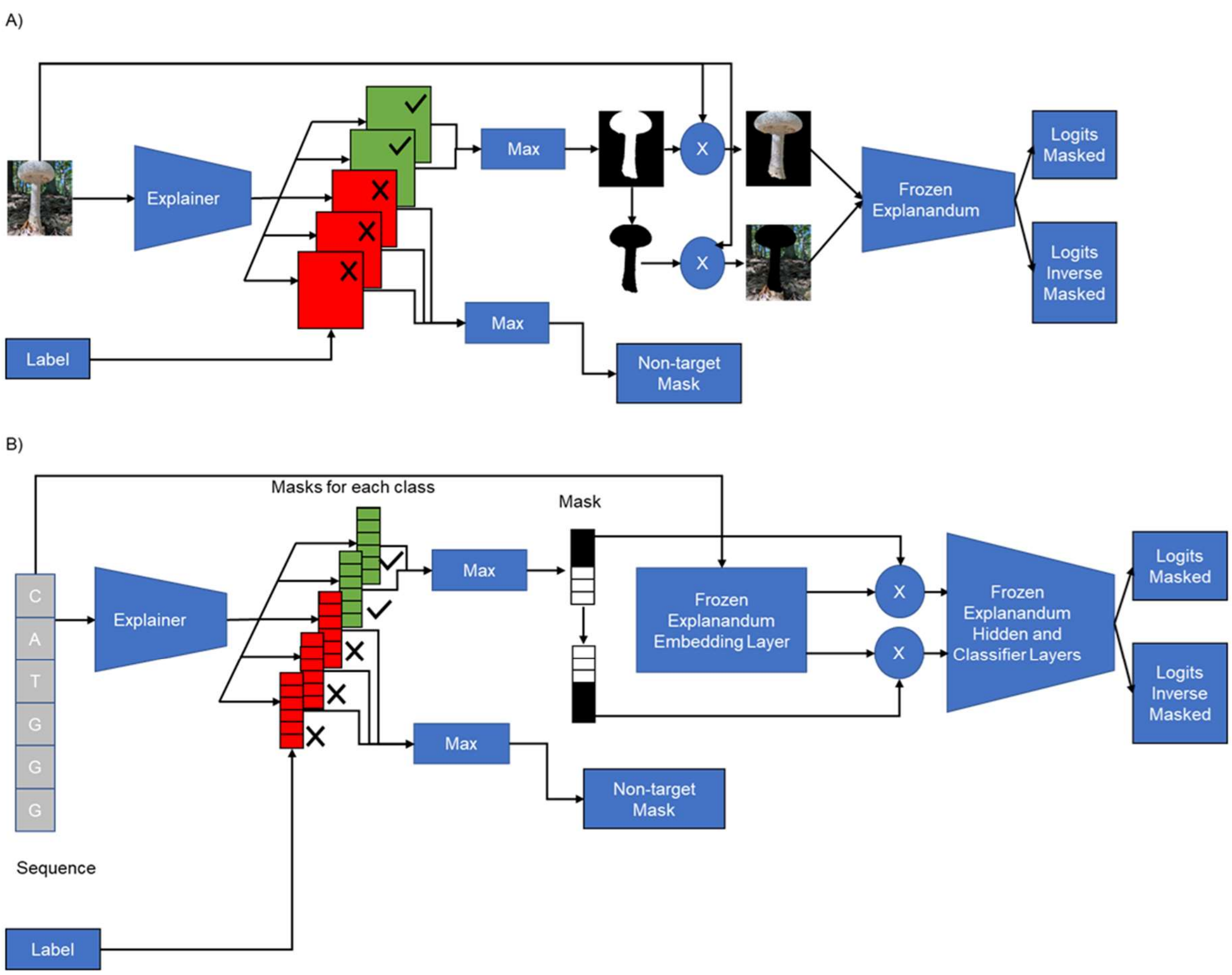

**Figure 1:** Overview of the Stalder et al. (11) approach for a) images (original paper) and b) tokenized inputs (our contribution). The fundamentals of the method are the same with the exception that there is a splitting of the Explanandum $\mathcal{F}$ model into a pre-embedding model and classification layers. The weights of the Explanandum in both cases are frozen and the weights of the Explainer $\mathcal{E}$ are trained to generate masks $s_c$ for each class. For the masks for the associated label, indicated by a ✓ (in green), the element-wise maximum is taken to make the target mask $m$ and its inverse, $\widetilde{m}$. This is done similarly for the masks not associated with the labels, indicated by a X (in red) to make the non-target mask $n$. For the case of a) the mask $m$ has the same dimensions as the input image, so the Hadamard product of the image and mask can be passed through the classifier. The same is done with the inverted mask. For the case of b) the mask vector is repeated, expanding it from a matrix of dimensions $d \; x \; 1$, where $d$ is the length of the sequence, to $d \; x \; h$, where $h$ is the dimension of the embedding vector. The Hadamard product of this repeated mask is then taken against the embedding vectors.

However, directly masking a token can only be done with a zero or a one: the product of a mask value of 0.25 and a discrete token isn't a valid operation. One way around this is to use a specific [mask] token represented by the zero vector. There are two problems with this. The first is that not all models are implemented with this already built into the model architecture, making it difficult to implement. The second problem is that, if using the Explainer-Explanandum paradigm, you would need some type of step function to force a zero or one, which can make learning difficult. We overcome this challenge this by changing where the Hadamard product is taken. By separating Explanandum into pre- and post-embedding layers, then taking the Hadamard product against the embedded tokens, we create the same result as using a [mask] token corresponding to the zero vector, without adding new tokens to the vocabulary or disrupting differentiability. This operation changes the magnitude of the embedding vector without changing the orientation of the embedding vector in the embedding space (**Figure 2**). Then, the masking operation is similar to that for an image, preserving differentiability.

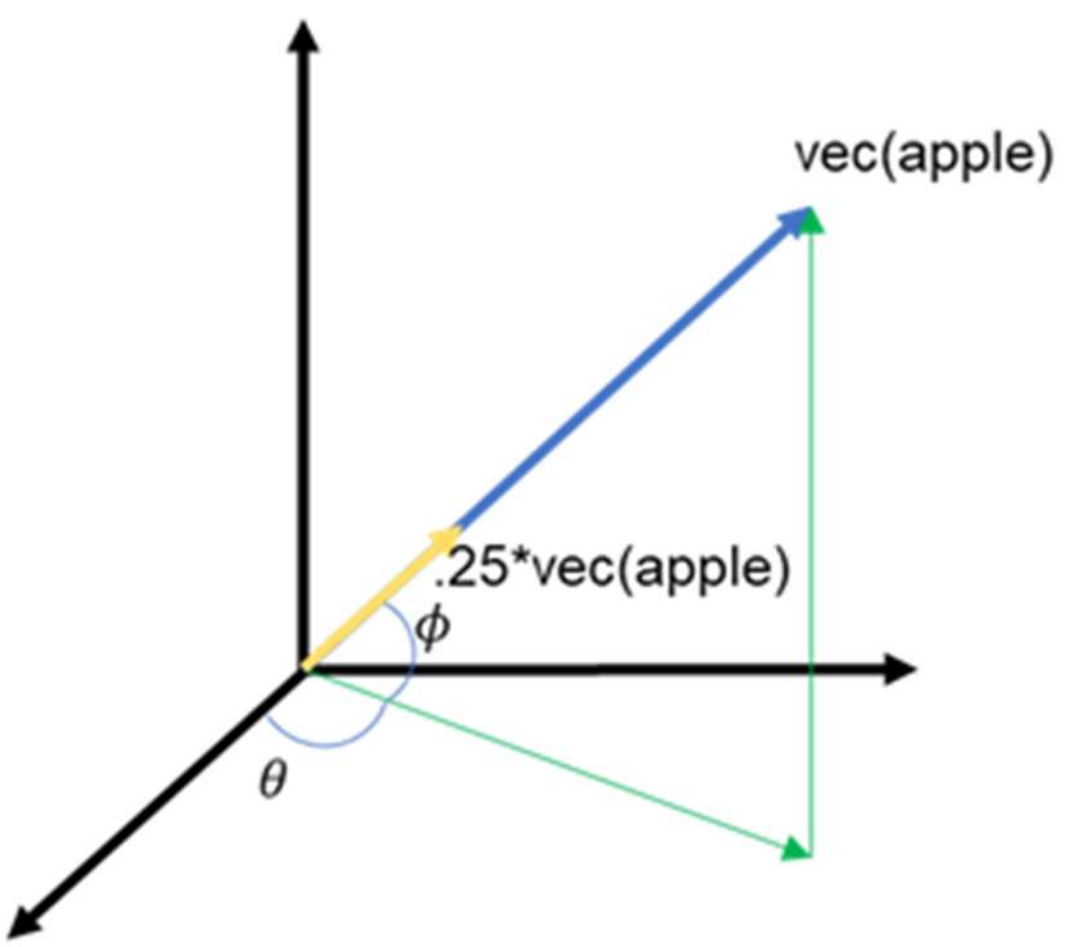

**Figure 2:** Demonstrates how an embedding vector (blue) is altered when elementwise multiplied against the mask (0.25). Because the mask vector is repeated, the same value is multiplied across the whole embedding vector. This dampens the magnitude of the original word vector for apple, giving it less impact, but the orientation is not changed (yellow). As a result, the fundamentals of the mathematics happening within the classifier are not impacted: the abstraction of the meaning of apple is preserved (it is not suddenly an orange), but its influence on the modeling task is diminished.

This approach works with any architecture, with one caveat: the softmax outputs for a blank sequence and for a sequence masked in its entirety should both be approximately uniform. Conceptually, this makes sense as inputting nothing should not result in any classification. We found that this was not always the case with all architectures, specifically in transformer architectures when the classifier uses a mean pooling across all the tokens, as opposed to just the [CLS] token output. With NT50m, mean pooling is taken for each token prior to the fully connected layers for classification, meaning an empty sequence or a completely masked sequence did not result in a uniform output of the Explanandum classifier. To fix this, we found it necessary to take the Hadamard product of the mask and the embedding and the Hadamard product of the mask and the output of the transformer prior to mean pooling and classification. The difference in architecture is shown in **Figure 3**. This is not necessary when using the CLS output token for classification.

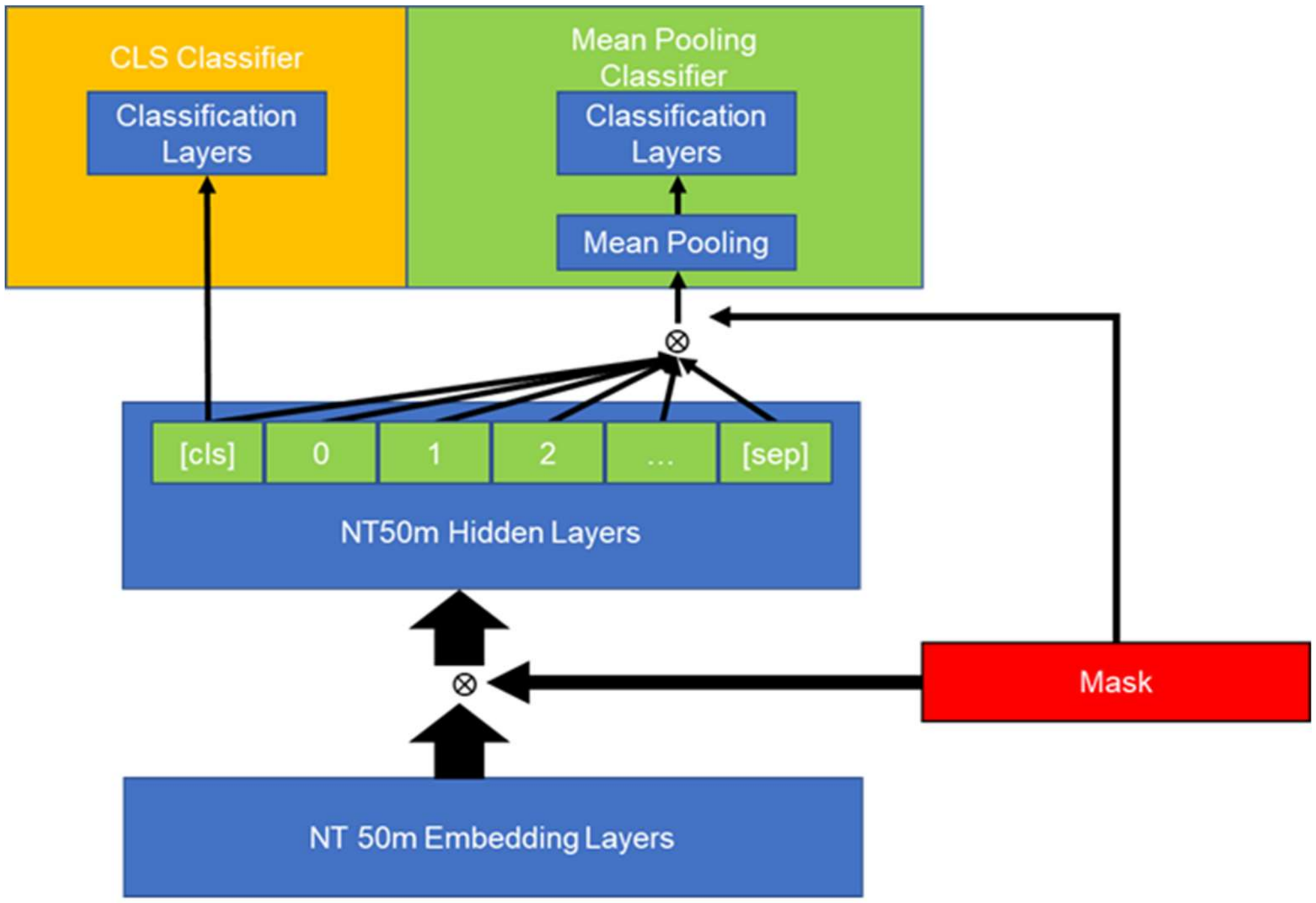

**Figure 3:** This figure shows the additional action that needed to be taken for a mean pooling classifier compared to a CLS classifier. Because there is a component from each token, the mask needs to be applied again to get good performance.

The second modification to adapt the what-you-see model to what-you-read functionality involves the loss function. This modification is the simple flattening of the formulation from two dimensions (for images) to one dimension (for text). More detail can be found in Stadler et al.(11). The new loss function has the same four components with the same hyperparameters:

$$L_E(\boldsymbol{x}, \mathcal{Y}, \boldsymbol{S}, \boldsymbol{m}, \boldsymbol{n}) = \boldsymbol{\mathcal{L}}_c(\boldsymbol{x}, \boldsymbol{\mathcal{Y}}, \boldsymbol{m}) + \lambda_e \boldsymbol{\mathcal{L}}_e(\boldsymbol{x}, \widetilde{\boldsymbol{m}}) + \lambda_a \boldsymbol{\mathcal{L}}_a(\boldsymbol{m}, \boldsymbol{n}, \boldsymbol{S}) + \lambda_{tv} \mathcal{L}_{tv}(\boldsymbol{m}, \boldsymbol{n}),$$

where $\boldsymbol{x}$ is the input sequence, $\mathcal{Y}$ is the class(es) the input $\boldsymbol{x}$ belongs to, $\boldsymbol{m}$ and $\widetilde{\boldsymbol{m}}$ are the mask and mask complement, and $\boldsymbol{n}$ is the masks for the non-correct classes. For our experiment, the hyperparameters $\lambda_e$, $\lambda_a$, and $\lambda_{tv}$ are all 1.

The loss term $\mathcal{L}_c$ is the standard classification loss using cross-entropy, to accommodate multi-class classification. The loss term $\mathcal{L}_e$ is the negative entropy of the complement of the mask $\boldsymbol{m}$. When the complement of the mask is applied, only the least important tokens are exposed; then, maximizing the entropy of the downstream classification probabilities is equivalent to maximizing the uncertainty, pushing the probabilities as near to uniform as possible when important tokens are masked. Text that has been stripped of relevant information should not produce a meaningful classification result.

$$\mathcal{L}_e = \frac{1}{C} \sum_{c=1}^{C} (\mathcal{F}(\boldsymbol{x} \odot \widetilde{\boldsymbol{m}})[c]) \log(\mathcal{F}(\boldsymbol{x} \odot \widetilde{\boldsymbol{m}})[c])$$

where $\mathcal{F}(\cdot)$ is the Explanandum and outputs a set of probabilities.

The loss term $\mathcal{L}_a$ is the area loss. As explained in Stadler, et al.(11), the first two loss terms do not directly impact the mask itself. The first loss term $\mathcal{L}_c$ incentivizes the model to create a mask that does not hide anything, while the second loss term $\mathcal{L}_e$ incentivizes the model to create a mask that hides everything. $\mathcal{L}_a$ introduces some balance, such that the mask neither obscures nor reveals too much. The first two summands in $\mathcal{L}_a$ are the mean values of the mask, $\mathcal{A}(\boldsymbol{m}) = \frac{1}{Z}\sum_i m[i]$, where $Z$ is the length of the sequence, and the mean values of the non-target mask, $\mathcal{A}(\boldsymbol{n}) = \frac{1}{Z}\sum_i n[i]$.

In addition, a valid, fractional minimum area $a$ and maximum area $b$ is prespecified, where $0 < a < b < 1$. Anything larger than $b$ covers too much area, and anything smaller than $a$ covers too little area. The third summand in $\mathcal{L}_a$ sorts the values of the class segmentation mask $s_c$, $r(s_c) = \boldsymbol{q}_{mask}$. Then we need to define two vectors $\boldsymbol{q}_{min}$ and $\boldsymbol{q}_{max}$ that contain 1s and 0s, with the number of 1s equal to a rounded $a * Z$ and $b * Z$, respectively. The bounding measure $\mathcal{B}$ is defined:

$$\mathcal{B}(s) = \frac{1}{Z}\sum_{i}^{Z} \max(\boldsymbol{q}_{min}[i] - \boldsymbol{q}_{mask}[i], 0) + \frac{1}{Z}\sum_{i}^{Z} \max(\boldsymbol{q}_{mask}[i] - \boldsymbol{q}_{max}[i], 0).$$

There is only a penalty if the mask area is smaller than $a$ or larger than $b$. Then, the loss term $\mathcal{L}_a$ is:

$$\mathcal{L}_a(\boldsymbol{m}, \boldsymbol{n}, \boldsymbol{S}) = \boldsymbol{\mathcal{A}}(\boldsymbol{m}) + \boldsymbol{\mathcal{A}}(\boldsymbol{n}) + \frac{\sum_{c=1}^{C} [\![ c \in \mathcal{Y} ]\!] \boldsymbol{\mathcal{B}}(s_c)}{\sum_{c=1}^{C} [\![ c \in \mathcal{Y} ]\!]}.$$

The fourth loss term $\mathcal{L}_{tv}$ encourages local smoothness amongst neighbors, represented through total variation loss on the target and non-target masks:

$$\mathcal{L}_{tv} = \frac{1}{Z}\sum_{i} |m[i] - m[i+1]| + \frac{1}{Z}\sum_{i} |n[i] - n[i+1]|$$

## Explainer

The architecture for the Explainer is relatively straightforward recurrent neural network architecture (**Figure 4**). As sequences are variable in length, the core of the Explainer model is a bidirectional Long Short-Term Memory (LSTM) layer with two layers and 40 nodes followed by an output rectified linear unit (ReLU) activation function. The input to the LSTM is an embedding layer with a vocabulary of 4107, equal to the vocabulary of the NT50m model when dimensionality is one hundred. The bidirectional LSTM layers produce a vector for every input token, each with length of 160 after the memory and cell states of the forward and backward passes are concatenated. Following batch normalization, a dense layer is applied with a sigmoid activation function and length equal to that of the number of classes, $c$. The final output shape for a sequence of length $d$ is a $d \ x \ c$ dimensional matrix.

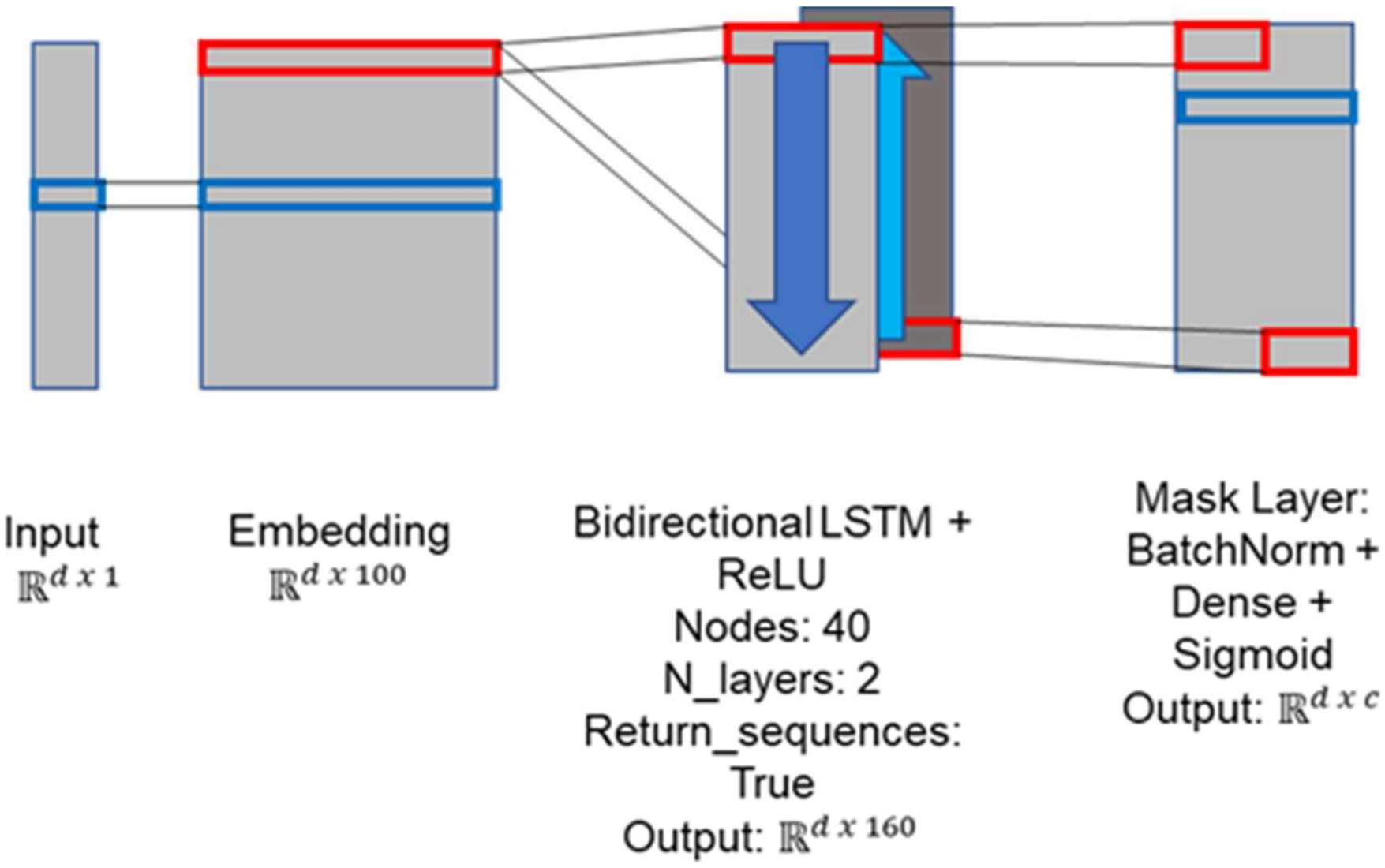

**Figure 4:** A diagram of the Explainer architecture, which consists of an embedding layer, a bidirectional LSTM with ReLU activation, a batch normalization layer and a dense layer with sigmoid activation for each class, resulting in a value ranging from 0 to 1 for each class and each token. The blue and red boxes at each layer are meant to demonstrate the flow of data through the network.

# Experiment

## Taxonomic data

The Explanandum in this experiment is a finetuned genomic language model (gLM) trained to perform taxonomic classification of DNA sequences, which we introduced in an earlier paper(14). The data was obtained from the open-access dataset published by Mock et al.(13), available as supplemental material: https://osf.io/dwkte(15). All sequences are 1500 base-pairs in length, standardized to all uppercase letters, and belong to four superkingdom categories: archaea, viruses, eukaryote, and bacteria. As the NT50m models use a 6-mer tokenizer (six amino acids grouped together, ex ACTGCC), the input sequences are expected to be 250 tokens, less than the maximum token length of NT50m of 1000 tokens. Phylum and genus labels are retrieved from the Environment for Tree Exploration (ETE) toolkit using the ncbi_taxonomy module. We use the December 2024 version of the publicly available taxdump archive (https://ftp.ncbi.nlm.nih.gov/pub/taxonomy/taxdump_archive/(16)). Sequences lacking either a phylum or genus label are excluded from our analysis.

The full processed dataset includes 5,181,880 sequences: 2,601,890 from Eukaryota, 1,828,018 from Bacteria, 524,276 from Archaea and 227,696 from Viruses. These are annotated with a total of 1,573 unique taxonomy IDs, spanning 4 superkingdoms, 55 phyla, and 1,878 genera. We reserve 2% of the full dataset stratified by genus labels (103,638 sequences) as a holdout test set. This test set is used to evaluate all experimental runs. We also reserve 10% of the full dataset to monitor model performance by evaluating validation loss over the course of training. The Explanandum was trained on the remainder of the dataset. The Explainer was trained on a subset of 239,391 examples randomly sampled from the training set, amounting to 4.7% of the training set, to reduce training time (20 minutes/epoch). The results on this subset were sufficient to render training on the full dataset unnecessary.

# Explanandum

The genomic language model NT50m is the backbone of the classifier. We utilize pretrained weights alongside a custom hierarchical classification layer. The classification layer consists of three classification heads, each a linear layer that maps the transformer's hidden size to the number of possible values for a given taxonomic rank: genus (1878 possibilities), phylum (55) and superkingdom (4). The forward pass sends token IDs and attention masks through the model backbone to produce token-level hidden states, which are then averaged using mean pooling across the sequences to get a fixed-length vector of length 512. Weighting this vector by the total number of valid (non-padded) tokens ensures that only meaningful tokens contribute to the mean embedding vector. Finally, the classification heads output a logit for the three taxonomic levels (the loss function is the sum of the three taxonomic level cross-entropies). An example of a masked sequence is shown in **Figure 5**.

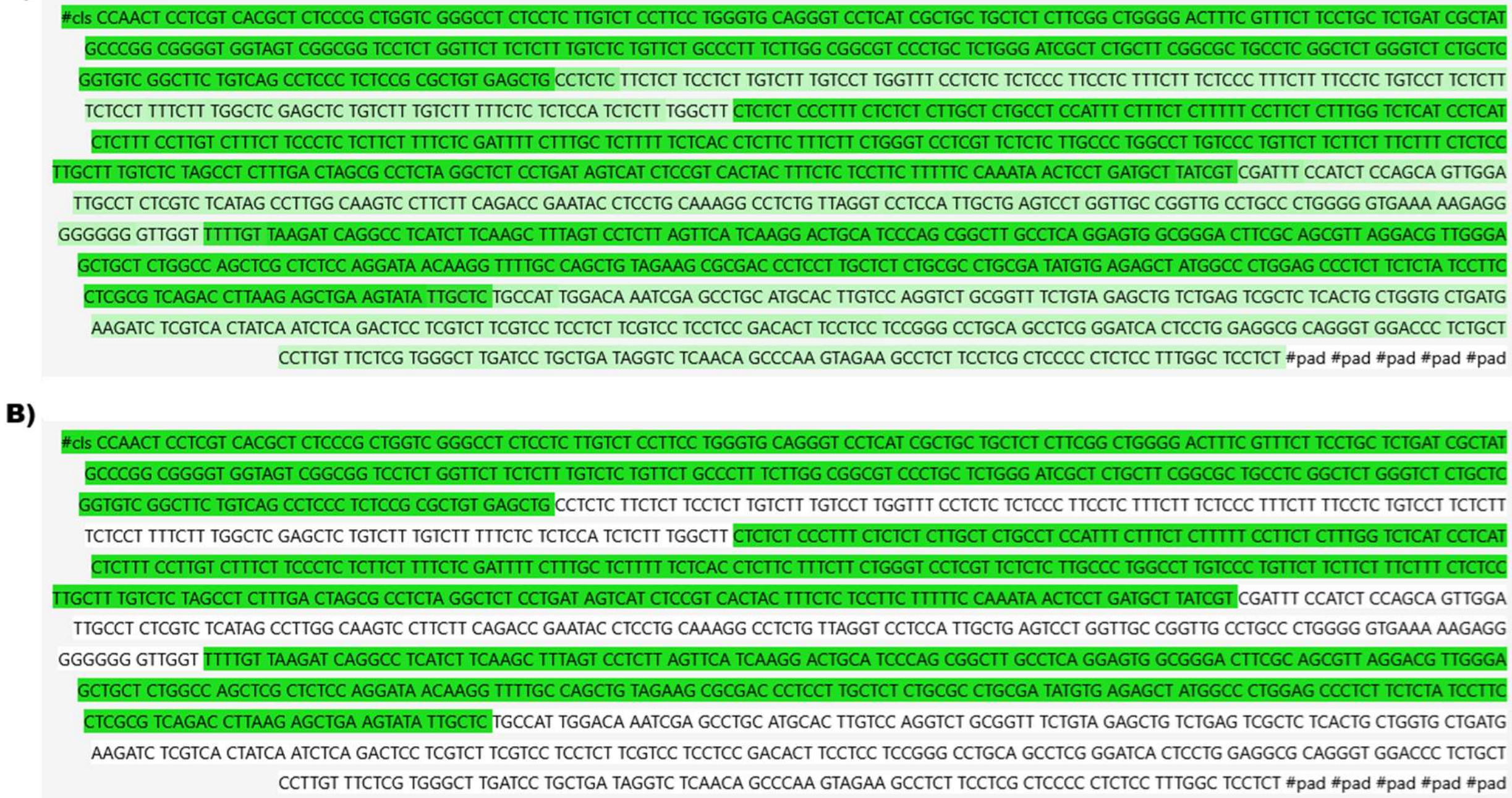

**Figure 5:** Examples of sequences correctly classified as from the Nyavirus genus. b) Importance values near 1 (very important) are represented by more intense green. Each set of six base-pairs is one token for NT50m. b) The same mask translation, rounded to either 0 or 1.

The Explanandum achieves a balanced classification accuracy of 99.05%, 97.39%, and 71.65% for superkingdom, phylum and genus, respectively, suggesting that there is a signal-rich feature set for the Explainer to find. We use balanced accuracy to ensure the weight of each taxonomic category is equally important.

# Training and resources:

Training of the Explainer and the Explanandum was done on a single NVIDIA H100 GPU with 80 GB of VRAM running Python 3.11.6 (17) with the PyTorch 2.5.1(18) library. Both models were trained with a learning rate of 0.0002 using a cosine learning rate scheduler to adjust the learning rate throughout training. contained batch sizes of 128 and 48, and a learning rate of 0.0002. The Explanandum was trained for five epochs with a batch size of 128. The Explainer was trained using a batch size of 48 and trained for 50 epochs, however, an early stopping mechanism ended training at 38 epochs. Training was conducted using FP32 precision, zero weight decay and no warm-up phase. Competing interpretability method LIME was implemented using the captum 0.7.0(19) Python package.

# Results

The Explainer successfully learned masks that cover the input sequences without compromising the classification performance of the Explanandum (**Table 1**). The results include performance on the unmasked sequences, the sequences with the masks, the rounded masks, the inverted masks, the inverted rounded masks, and finally, separated chunks. For the separated chunks, we took the rounded masks and separated the sequence into subsequences based continuous ones and zeros. Each chunk was classified individually and then the class was aggregated based on an average for the important and not important subsequences. This was done to demonstrate the residual effects of masked spacing were not impacting the classification significantly.

**Table 1:** Balanced accuracies of superkingdom, phylum and genus classification under various masking conditions, including the masks themselves, the mask complements, rounded masks, the inverted of the rounded masks, and separated chunks.

| | **Balanced Accuracy Superkingdom** | **Balanced Accuracy Phylum** | **Balanced Accuracy Genus** |
|---|---|---|---|
| Explanandum | 99.05% | 97.39% | 71.67% |
| Masks | 98.39% | 95.51% | 69.88% |
| Inverted Masks | 46.70% | 28.21% | 5.71% |
| Rounded Masks | 76.42% | 52.6% | 40.86% |
| Inverted Rounded Masks | 44.85% | 26.20% | 4.92% |
| Average relevant chunk | 69.98% | 49.92% | 36.32% |
| Average not-relevant chunk | 47.13% | 31.68% | 9.58% |

The unmasked classifier achieves balanced accuracies of 99.05%, 97.39%, and 71.65% for superkingdom, phylum and genus labels, respectively, while the masked classifier has comparable accuracies of 98.39%, 95.51%, and 69.88%. We confirm that the masks do not achieve this accuracy by simply exposing the entirety of all sequences: 67.28% of mask values over all tokens were greater than 0.5.

While there is a variation in the values of the mask at any given sequence, there is a consistent trend in the mask values, with many being close to 1 and a distribution spread out among the lower values, not quite getting to zero (**Figure 6a**), and there are clearly sections of the sequences, likely due to preprocessing, that are more likely to have a value closer to one (**Figure 6**b). The mean mask value is 0.73 with 67.28% greater than 0.5. However, a look at the maximum and minimum mask values at each location across test sequences indicates a variation in the mask values. The vast majority of sequence masks have three chunks of unmasked sequences, with a mean of 3.35 and about 90% having four or fewer, as shown in **Figure 7a**. The mean unmasked length was 51.39 tokens, with some exceeding 200 tokens, as shown in **Figure 7**Error! Reference source not found.**b**.

Checking both the inverted masks and the rounded masks confirm that the masks are removing relevant information. When the masks are inverted, accuracy of superkingdom, phylum and genus classification plummets from 98.39% to 46.70%, 95.51% to 28.21%, and 69.88% to 5.71%, respectively. The rounded masks show a drop, but nowhere near as steep, with superkingdom dropping to 76.42%, phylum to 52.6% and genus to 40.86%. Inverting the rounded masks gave similar results to the inverted masks, with balanced accuracies of 44.85%, 26.20% and 4.92%. All told, the masks effectively expose relevant information for classification while obscuring irrelevant information.

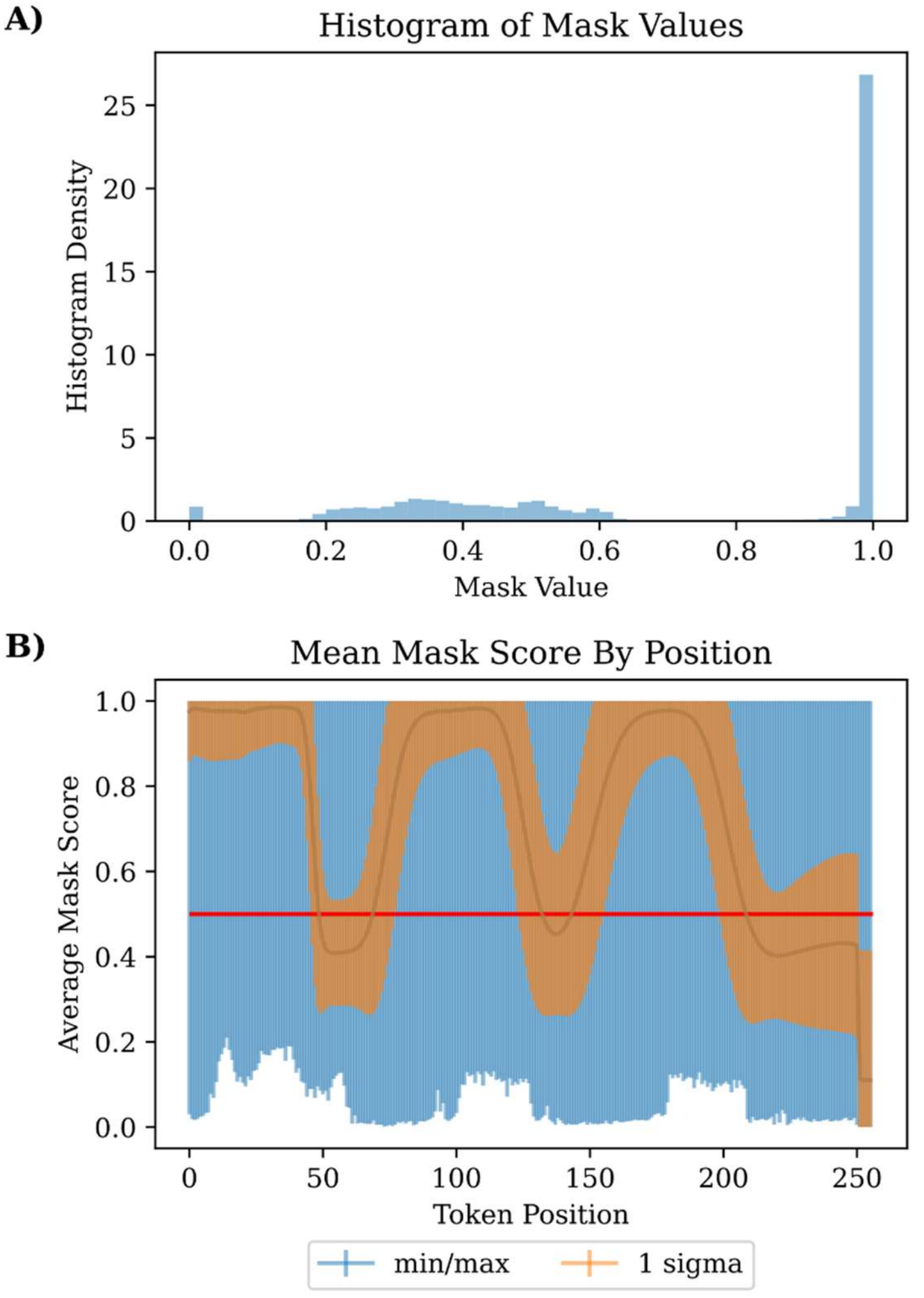

**Figure 6:** a) A histogram of the mask values across all test sequences. b) Average mask score by location in the sequence. The orange shading covers one standard deviation about the average value at a given position, while the blue shading shows the minimum and maximum range of values at a given position.

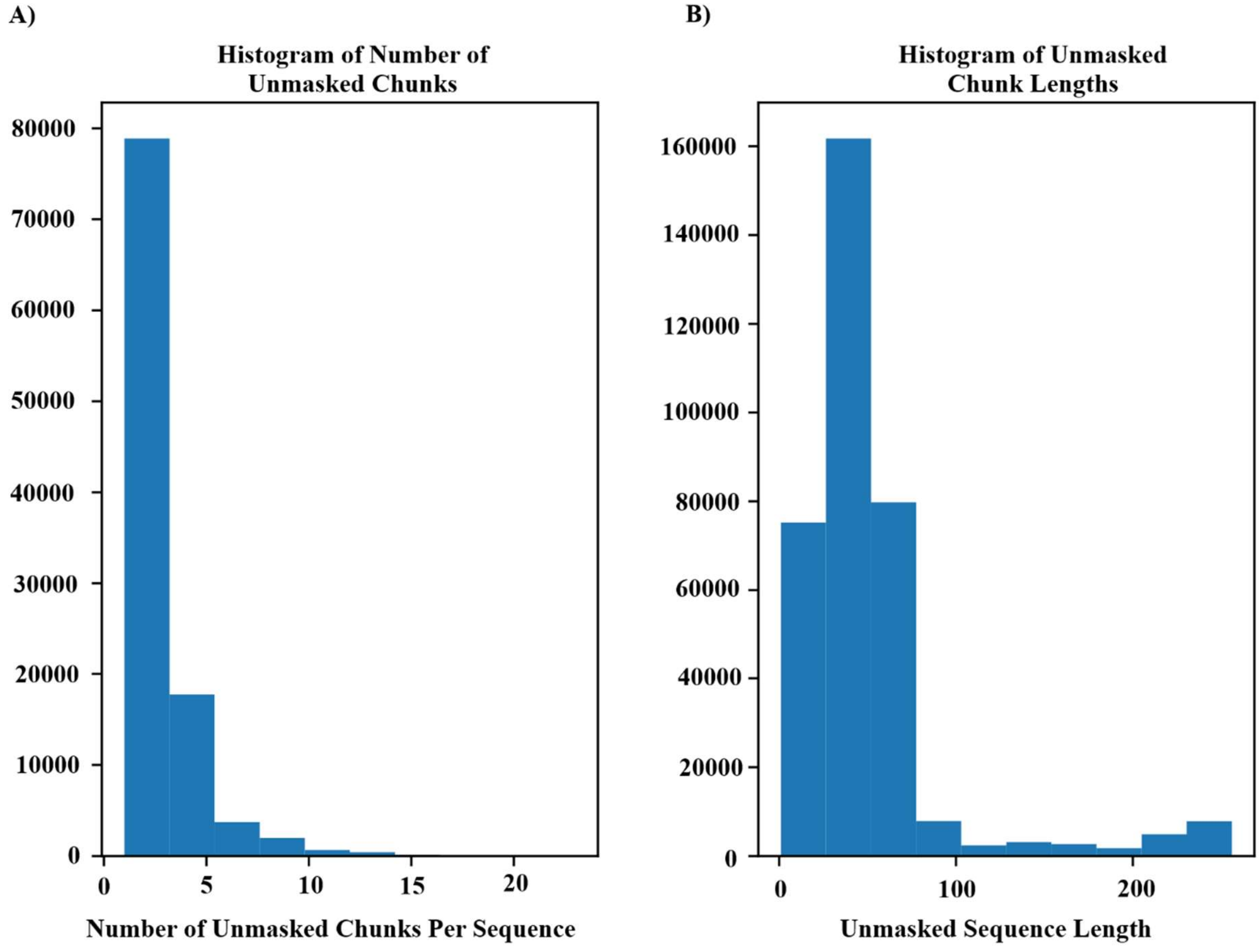

**Figure 7**: a) The distribution of the number of unmasked chunks after rounding the values to either 0 or 1, and b) the distribution of the lengths of unmasked chunks.

To confirm that blanks and masked out sequence segments do not play an influential role in classification, the sequences were broken into relevant and irrelevant chunks, which were then passed through the Explanandum. The relevant chunks had balanced classification accuracies of 69.98%, 49.92% and 36.32%, versus the irrelevant chunks' 47.13%, 31.68% and 9.58%. These performance metrics are similar to that of the masked and inverted-masked sequences.

Due to the emphasis on globality that underpins transformer-based architectures, the most popular explainable AI frameworks for text—LIME and Shap—are inherently ill-equipped to clarify taxonomic sequence classification decisions. This, combined with the non-serializable architecture of the Evolutionary Scale Modeling (ESM)-based backbone of NT50m, made the Python packages for LIME, transformer-interpret and Shap prohibitively difficult to use. Runtime is an additional limiting factor. LIME took an average of 26 seconds per-sequence, per-taxonomic rank, which would amount to months if run without parallelization on the 103,638 sequence testing set. Compared to our variant on Stadler et al.(11), the Explainer was trained in 15 hours and 40 minutes, and took twenty minutes to generate masks for the 103,638 testing set sequences. Even if it was feasible to run LIME at scale, the outputs would not be reliably helpful. **Figure 8** demonstrates the output of LIME on an example at the phylum level that was correctly classified correctly. From the image, it's unclear which of the green chunks (positively contributed to classification) were most significant and the presence of red (negatively contributing to classification) only create additional confusion. As you filter down to show only the ones that contributed positively and even only substantially (>0.5) the tokens are very much spread out, providing little interpretability. Meanwhile, the masks demonstrated in **Figure 8** are substantially more interpretable.

A) Example of LIME output

ATAAAA AAAAAC CAATAC ATTATT TTAGCA AAACAA CAAAAG ACTCAC TTTAGC TGGGTT CCTGAT TCAGCA TGCCGC TTAAGG AGCTGC CGAGTC ACTCAA GCGCTT ATAGCA TGCTGG CTTCTT CATGCT CATCCG
CACATG GCTGTG ACTCAA AGTGAC TGGATA CTGTAT GCAACA TACAGT GTAAAA ATCCAG TCAGTC ATCCAA ACAACT CATATC ACTATA TACTGT GTATGT AGTGTG TAAGGG CATTAC TGTGTG GATATG AGTTGG
TCAGGG GCTTCA AAGAGC TTTCAG TTGCTT ATTATT GAGCAT CTCAGC TGGGAT TAGGAT TAGCTG GTTTAT TCCTCA GTTAGT TGGAAA AATACC AGCACC TGGAAT TTCCAG GACAAT CATTCC CATGTG GATATA
CATTAA AGATGT TGGTCT TGTTTT TTTTTT TTTTAG TTTCCA TTAAAT ATTAAA TGCTTC TCAAAC TATAAC CAGTTA CATGAG TTTGAA GTCATT TTTTCC TCTCTT TATCTG ATAGTT TGCACG TTTGAG CACTAA AAGATT
GTGCAC ATCAAA GCACCT GCATTT CCTGAA TATCCT TAACAA AACAAA CCCAGT CAACCT CCTGTC ACATAC CTGCTC TCCCAC CCTTTT TCACAG TCAGAA TTACCC AAAAAC TTCAAA TTTAGT GATTTT GTTTGG
TGAAAT CTGTTC GTTTCT GGGGCT TAGCTA AAGGTT TTATCT GTTTAA TGATGC ATGTCT CCTTCT GAAGGT GTATCC CTATGA AATGCT CATGAT AACCAG TAGAGG GAGAGC TAAACT GCCAAG AGATGT GGACAG
AACCAG ACTGGA GGTATC AGTTTT CCTCCT GACGCT CCTCAG TTAACA CTGCAT TTTGAT TTTGTG TCAACA GCATTG TTACAC AAACTT ACATCT GTAGCT GAGAAC ATGCAG CATCAG TTGCTC CTTGTC ATTTTT
GCAGCA AACCCC TTTTTT TACAGC TTTGAT GCTAAC TGAGAA CCCCCC ACCTTC CCCCCT CCTTTG TTTTTT CTCCAC CTTTGT TCTCAT TTCTAG CGCCAC TTAGCA CCCGAA ACGTTC TTTGAG ATCTTT GGAATG
GAGATC CAGGAA TTTGAC AGGCTT CCCCTG TGGAAA CGCAAC GACATG AAAAAG AAGGCC AAGCTC TTTTAG TTGTTG TTGTTT TTTTTG TTTTTT TTTTCT AATTTA TAGCAT GCATTA TTCTGT ACATAC AATCAG
ATAGCA CACAGT TGTAAT CTAGAT GTAGGT TCTGGG TTATTC CTTTTC TTTTTT TTTTTT TTTTTT TTTGCT TATGTC ACCTGG ATATAC CTGGGG ATGCCA AGAAAT GGACTG ATGTTG CGATGA TGGACT TCAATA
ATGAAT GGAAAC CTCTTC CATGTT CAACAA AGAATG ATGAGA AGGGTC AAGTTT TTTTTG CTTTGA CTGCAC TTATTT TACCTC CTGCTT TGCTTT AAGTGA ATCCTT CCTTTA

B) Example of LIME output with zero threshold

ATAAAA AAAAAC CAATAC ATTATT TTAGCA AAACAA CAAAAG ACTCAC TTTAGC TGGGTT CCTGAT TCAGCA TGCCGC TTAAGG AGCTGC CGAGTC ACTCAA GCGCTT ATAGCA TGCTGG CTTCTT CATGCT CATCCG
CACATG GCTGTG ACTCAA AGTGAC TGGATA CTGTAT GCAACA TACAGT GTAAAA ATCCAG TCAGTC ATCCAA ACAACT CATATC ACTATA TACTGT GTATGT AGTGTG TAAGGG CATTAC TGTGTG GATATG AGTTGG
TCAGGG GCTTCA AAGAGC TTTCAG TTGCTT ATTATT GAGCAT CTCAGC TGGGAT TAGGAT TAGCTG GTTTAT TCCTCA GTTAGT TGGAAA AATACC AGCACC TGGAAT TTCCAG GACAAT CATTCC CATGTG GATATA
CATTAA AGATGT TGGTCT TGTTTT TTTTTT TTTTAG TTTCCA TTAAAT ATTAAA TGCTTC TCAAAC TATAAC CAGTTA CATGAG TTTGAA GTCATT TTTTCC TCTCTT TATCTG ATAGTT TGCACG TTTGAG CACTAA AAGATT
GTGCAC ATCAAA GCACCT GCATTT CCTGAA TATCCT TAACAA AACAAA CCCAGT CAACCT CCTGTC ACATAC CTGCTC TCCCAC CCTTTT TCACAG TCAGAA TTACCC AAAAAC TTCAAA TTTAGT GATTTT GTTTGG
TGAAAT CTGTTC GTTTCT GGGGCT TAGCTA AAGGTT TTATCT GTTTAA TGATGC ATGTCT CCTTCT GAAGGT GTATCC CTATGA AATGCT CATGAT AACCAG TAGAGG GAGAGC TAAACT GCCAAG AGATGT GGACAG
AACCAG ACTGGA GGTATC AGTTTT CCTCCT GACGCT CCTCAG TTAACA CTGCAT TTTGAT TTTGTG TCAACA GCATTG TTACAC AAACTT ACATCT GTAGCT GAGAAC ATGCAG CATCAG TTGCTC CTTGTC ATTTTT
GCAGCA AACCCC TTTTTT TACAGC TTTGAT GCTAAC TGAGAA CCCCCC ACCTTC CCCCCT CCTTTG TTTTTT CTCCAC CTTTGT TCTCAT TTCTAG CGCCAC TTAGCA CCCGAA ACGTTC TTTGAG ATCTTT GGAATG
GAGATC CAGGAA TTTGAC AGGCTT CCCCTG TGGAAA CGCAAC GACATG AAAAAG AAGGCC AAGCTC TTTTAG TTGTTG TTGTTT TTTTTG TTTTTT TTTTCT AATTTA TAGCAT GCATTA TTCTGT ACATAC AATCAG
ATAGCA CACAGT TGTAAT CTAGAT GTAGGT TCTGGG TTATTC CTTTTC TTTTTT TTTTTT TTTTTT TTTGCT TATGTC ACCTGG ATATAC CTGGGG ATGCCA AGAAAT GGACTG ATGTTG CGATGA TGGACT TCAATA
ATGAAT GGAAAC CTCTTC CATGTT CAACAA AGAATG ATGAGA AGGGTC AAGTTT TTTTTG CTTTGA CTGCAC TTATTT TACCTC CTGCTT TGCTTT AAGTGA ATCCTT CCTTTA

C) Example of LIME output with 0.5 threshold

ATAAAA AAAAAC CAATAC ATTATT TTAGCA AAACAA CAAAAG ACTCAC TTTAGC TGGGTT CCTGAT TCAGCA TGCCGC TTAAGG AGCTGC CGAGTC ACTCAA GCGCTT ATAGCA TGCTGG CTTCTT CATGCT CATCCG
CACATG GCTGTG ACTCAA AGTGAC TGGATA CTGTAT GCAACA TACAGT GTAAAA ATCCAG TCAGTC ATCCAA ACAACT CATATC ACTATA TACTGT GTATGT AGTGTG TAAGGG CATTAC TGTGTG GATATG AGTTGG
TCAGGG GCTTCA AAGAGC TTTCAG TTGCTT ATTATT GAGCAT CTCAGC TGGGAT TAGGAT TAGCTG GTTTAT TCCTCA GTTAGT TGGAAA AATACC AGCACC TGGAAT TTCCAG GACAAT CATTCC CATGTG GATATA
CATTAA AGATGT TGGTCT TGTTTT TTTTTT TTTTAG TTTCCA TTAAAT ATTAAA TGCTTC TCAAAC TATAAC CAGTTA CATGAG TTTGAA GTCATT TTTTCC TCTCTT TATCTG ATAGTT TGCACG TTTGAG CACTAA AAGATT
GTGCAC ATCAAA GCACCT GCATTT CCTGAA TATCCT TAACAA AACAAA CCCAGT CAACCT CCTGTC ACATAC CTGCTC TCCCAC CCTTTT TCACAG TCAGAA TTACCC AAAAAC TTCAAA TTTAGT GATTTT GTTTGG
TGAAAT CTGTTC GTTTCT GGGGCT TAGCTA AAGGTT TTATCT GTTTAA TGATGC ATGTCT CCTTCT GAAGGT GTATCC CTATGA AATGCT CATGAT AACCAG TAGAGG GAGAGC TAAACT GCCAAG AGATGT GGACAG
AACCAG ACTGGA GGTATC AGTTTT CCTCCT GACGCT CCTCAG TTAACA CTGCAT TTTGAT TTTGTG TCAACA GCATTG TTACAC AAACTT ACATCT GTAGCT GAGAAC ATGCAG CATCAG TTGCTC CTTGTC ATTTTT
GCAGCA AACCCC TTTTTT TACAGC TTTGAT GCTAAC TGAGAA CCCCCC ACCTTC CCCCCT CCTTTG TTTTTT CTCCAC CTTTGT TCTCAT TTCTAG CGCCAC TTAGCA CCCGAA ACGTTC TTTGAG ATCTTT GGAATG
GAGATC CAGGAA TTTGAC AGGCTT CCCCTG TGGAAA CGCAAC GACATG AAAAAG AAGGCC AAGCTC TTTTAG TTGTTG TTGTTT TTTTTG TTTTTT TTTTCT AATTTA TAGCAT GCATTA TTCTGT ACATAC AATCAG
ATAGCA CACAGT TGTAAT CTAGAT GTAGGT TCTGGG TTATTC CTTTTC TTTTTT TTTTTT TTTTTT TTTGCT TATGTC ACCTGG ATATAC CTGGGG ATGCCA AGAAAT GGACTG ATGTTG CGATGA TGGACT TCAATA
ATGAAT GGAAAC CTCTTC CATGTT CAACAA AGAATG ATGAGA AGGGTC AAGTTT TTTTTG CTTTGA CTGCAC TTATTT TACCTC CTGCTT TGCTTT AAGTGA ATCCTT CCTTTA

**Figure 8:** Example of LIME outputs for a correctly classified example where a) shows the scores out of LIME ranging from 1 (green, positive importance) to -1 (red, negative importance). Color saturation represents magnitude of the result. b) The same values with a lower bound of 0, and c) the same values with a lower bound of 0.5.

# Conclusions

We introduced an extension of a proven explainable AI method for images to text by learning a continuous-scale mask on the embeddings, avoiding the mathematical pitfall of a discrete mask on the tokens. This approach utilizes a deep learning model to create a mask that is repeated and applied to each embedding vector using the Hadamard product. We demonstrated its utility on an NT50m model trained for multi-level taxonomic classification, using only a fraction of the data that had been used to train the NT50m model. The Explanandum achieved superior classification accuracy when regions deemed unimportant to classification by the Explainer are obscured, compared to when regions deemed important to classification by the Explainer are obscured. This holds even when mask values are rounded to be more extreme (and less precise). In addition, the time required to train and apply the Explainer is substantially faster than methods that do not require continuity. As a result, this method is well-suited to explaining classification decisions on sequences, and can easily be generalized to text. Future work includes applications to text or protein or genomic functional space, to improve understanding of how advanced transformer-based models work.

# Code Availability

Model training and inference code used in this study is available at [https://github.com/jhuapl-bio/What_You_Read_ExplainableAI](https://github.com/jhuapl-bio/What_You_Read_ExplainableAI)

# Acknowledgements

We would like to thank Dr. Molly Gallagher for valuable contributions and for helping us navigate programmatic hurdles during the completion of this work.

# Funding

This work was supported by funding from the U.S. Centers for Disease Control and Prevention through the Office of Readiness and Response under Contract # 75D30124C20202.